\pdfoutput=1 
\RequirePackage{fix-cm}
\documentclass[twocolumn]{svjour3}          

\usepackage[]{natbib}
\usepackage{amsmath,graphicx}
\usepackage[]{hyperref}
\hypersetup{pdftitle={Validation of Tsallis Entropy In Inter-Modality Neuroimage Registration},
colorlinks=true,citecolor=blue,linkcolor=red}

\begin{document}

\title{Validation of Tsallis Entropy In Inter-Modality Neuroimage Registration\thanks{Funding Support:  ÒS\~{a}o Paulo Research FoundationÓ -- FAPESP, Grants\# 2010/10979-6 and 2013/08240-0 (Henrique Tomaz Amaral-Silva).}
}


\author{Henrique Tomaz Amaral-Silva \and
        Luiz Otavio Murta-Jr \and
        Paulo Mazzoncini de Azevedo-Marques \and
        Lauro Wichert-Ana \and
        V. B. Surya Prasath \and
        Colin Studholme
}


\institute{H. T. Amaral-Silva \at
              Biomedical Image Computing Group, Departments of Pediatrics and Bioengineering, University of Washington, Seattle, USA. Also with Center of Imaging Sciences and Medical Physics, Internal Medicine Department and Graduate Program, Ribeir\~{a}o Preto Medical School, University of S‹o Paulo, Ribeir\~{a}o Preto, Brazil\\
              \email{amaralht@uw.edu}         \\
             \emph{Present address:}
             Siemens Healthineers, S\~{a}o Paulo, Brazil
           \and
           L. Otavio Murta-Jr \at
             Department of Computing and Mathematics, University of S\~{a}o Paulo, Ribeir\~{a}o Preto, Brazil.
             \email{murta@ffclrp.usp.br}
             \and
            P. M. Azevedo-Marques \at
            Center of Imaging Sciences and Medical Physics, Internal Medicine Department and Graduate Program, Ribeir\~{a}o Preto Medical School, University of S‹o Paulo, Ribeir\~{a}o Preto, Brazil. Also with Bioengineering Interunits Graduate Program, S‹o Carlos School of Engineering, University of S\~{a}o Paulo, S\~{a}o Carlos, Brazil. 
            \email{pmarques@fmrp.usp.br}
            \and
            L. Wichert-Ana \at
            Center of Imaging Sciences and Medical Physics, Internal Medicine Department and Graduate Program, Ribeir\~{a}o Preto Medical School, University of S\~{a}o Paulo, Ribeir\~{a}o Preto, Brazil. Also with Bioengineering Interunits Graduate Program, S\~{a}o Carlos School of Engineering, University of S\~{a}o Paulo, S\~{a}o Carlos, Brazil, and Nuclear Medicine Section, Ribeir‹o Preto Medical School, University of S‹o Paulo, Ribeir‹o Preto, Brazil. 
            \email{lwichert@fmrp.usp.br}
            \and
            V. B. Surya Prasath \at
            Computational Imaging and VisAnalysis (CIVA) Lab, Department of Computer Science, University of Missouri-Columbia, MO 65211 USA.
            \email{prasaths@missouri.edu}
            \and
            C. Studholme \at
            Departments of Pediatrics, Bioengineering, and Radiology, University of Washington, Seattle, USA.
            \email{studholm@uw.edu}
}

\date{Received: date / Accepted: date}

\maketitle

\begin{abstract}
Medical image registration plays an important role in determining topographic and morphological changes for functional diagnostic and therapeutic purposes. Manual alignment and semi-automated software still have been used; however they are subjective and make specialists spend precious time. Fully automated methods are faster and user-independent, but the critical point is registration reliability. Similarity measurement using Mutual Information (MI) with Shannon entropy (MIS) is the most common automated method that is being currently applied in medical images, although more reliable algorithms have been proposed over the last decade, suggesting improvements and different entropies; such as~\cite{Studholme1999}, who demonstrated that the normalization of Mutual Information (NMI) provides an invariant entropy measure for 3D medical image registration. In this paper, we described a set of experiments to evaluate the applicability of Tsallis entropy in the Mutual Information (MIT) and in the Normalized Mutual Information (NMIT) as cost functions for Magnetic  Resonance Imaging (MRI), Positron Emission Tomography (PET) and Computed Tomography (CT) exams registration. The effect of changing overlap in a simple image model and clinical experiments on current entropies (Entropy Correlation Coefficient - ECC, MIS and NMI) and the proposed ones (MIT and NMT) showed NMI and NMIT with Tsallis parameter close to 1 as the best options (confidence and accuracy) for CT to MRI and PET to MRI automatic neuroimaging registration.
\keywords{registration \and Tsallis entropy \and inter-modality \and neuroimaging.}
\end{abstract}

\section{Introduction}
Systems for Computer Aided Diagnosis (CAD) have played a significant role as an auxiliary tool in medical diagnosis~\citep{Doi2007}. In particular, CAD may contribute to the visualization, interpretation and quantification of medical images. Several processing techniques may be applied to image segmentation, smoothing, realignment, normalization and registration~\citep{Doi2007,Azevedo-Marques2003}. The registration process consists of overlaying two or more images using the coordinate points in one image to establish the correspondence to the coordinate points in the other one. This process can be performed by three ways: (a) manually by a specialist who needs to drive the reference image to the target one, (b) using semi-automated registration, based on markers defined in the own images, and (c) using automated registration, based on algorithms which generally use intrinsic characteristics of the image, such as surfaces, contours and voxel intensity values~\citep{Hajnal2001,diez2014intensity}. These algorithms can perform non-rigid registration, deforming the original image with scales and shear transformations (used to inter-subject registration or to normalize an image to a standardized space Ð such as an atlas). However, rigid body registration is the most commonly used method, once it is applied to intra-subject image registration that consists of finding rotations and translations to match images with same anatomical shape; for instance, a 3D-to-3D rigid body registration is defined by six parameters: three translations and three rotations~\citep{Ashburner2004}. 

On clinical grounds, the findings obtained by different imaging techniques can be used in a complementary manner by physicians. By combining morphologic (X-ray, Ultrasound (US), Computed Tomography (CT) and Magnetic Resonance Imaging (MRI)) and functional (Positron Emission Tomography (PET), Single Photon Emission Computed Tomography (SPECT) and functional Magnetic Resonance Imaging (fMRI)), data can improve image interpretation~\citep{Gianfranco2008}. This fusion of images, known as multimodal or intermodal medical image registration, is an important tool for surgical planning. In neurosurgery, for example, multimodal or intermodal image registration is useful to identify lesions or to localize dysfunctional areas to be resected. For epilepsy treatment, registration of structural and functional images can improve the localization of the epileptogenic zone that must be surgically removed for patients to become seizure-free~\citep{McNally2005}. Monomodal or intramodal registration has important applications in the comparison of pre and post-condition images, eg., comparison of ictal and interictal SPECT images~\citep{mumcuoglu2006simultaneous}, subtraction of ictal and interictal SPECT co-registered to MRI (SISCOM)~\citep{fuster2013focusdet}, and for the study of ongoing or stepwise brain changes in the course of a disease~\citep{liao2012statistical,tang2016groupwise}, see \citep{viergever2016survey} for a recent review.

Software for neuroimage processing generally uses techniques based on voxel intensity values for automatic registration.  They might be based in the following cost functions: (1) Mutual Information using Shannon entropy (MIS), (2) Entropy of Correlation Coefficient (ECC), (3) Normalized Mutual Information (NMI) and (4) Normalized Cross Correlation (NCC). MIS is a measurement originated from Information Theory~\citep{Shannon1948}, and was first proposed for medical image registration by~\cite{Collignon1995} and~\cite{Viola1997}. Since then, several studies have found satisfactory results for registration based on Information Theory. Recently, a comparative study of 16 registration methods evaluated rigid body registration performance using multimodal neuroimages. The results have shown that two methods based on MI were superior to the alternative ones and obtained similar accuracy to the gold standard~\citep{Gao2008}. 
 
The image entropy can be interpreted as a way to measure the dispersion of a gray level distribution. Homogeneous images have low entropy value when the gray level distribution has only one threshold. High contrast images, with many grey levels, result in a high entropy value~\citep{Pluim2000}. In this setting, given two images $A$ and $B$, the definition of Mutual Information is:
\begin{eqnarray}\label{E:MI}
MI(A,B) = H(A) + H(B) - H(A,B)
\end{eqnarray}
where $H(A)$ and $H(B)$ and  correspond to the entropy of images $A$  and $B$, respectively, and $H(A,B)$  to the intersection of the entropies, corresponding to a measure of the dispersion with the joint distribution probability $p(a,b)$. 
In other words, $H(A,B)$ is the probability of incidence of the gray value $a$ in image $A$ and the grey value $b$ in image $B$ (in the same position), for all $a$ and $b$ in overlapped region of $A$ and $B$.

The classic Shannon entropy is represented by the formula,
\begin{eqnarray}\label{E:shannon}
H = -k\,\sum_{i=1}^W p_i \log{p_i}.
\end{eqnarray}
This formalism was demonstrated to be restricted to the Boltzmann-Gibbs-Shannon (BGS) domain, which is assumed to adequately describe the behavior of a system when the effect of interactions and the microscopic memory are short-ranged. Usually, systems that conform to BGS are referred to as additive systems. Considering that a physical system can be decomposed into two statistically independent subsystems $A$ and $B$, the probability of a composite system is:
\begin{eqnarray}\label{E:composite}
p^{A+B} = p^A * p^B
\end{eqnarray}
thus verifying that the Shannon entropy has the additive property:
\begin{eqnarray}\label{E:shannonadditive}
S(A+B) = S(A) + S(B)
\end{eqnarray}
However, for certain classes of physical systems that possess long-range interactions, long-time memories and fractal-type structure, it is necessary to extend the classical theory presented by Shannon. The relevance of fractal geometry in medical image processing is explained by the self-similarity observed in biological structures imaged with a finite resolution. These images are not only spectrally and spatially complex, but also exhibit similarities at different spatial scales~\citep{Lopes2009}.  Inspired 
by multifractal concept,~\cite{Tsallis1988} proposed a generalization of BGS statistics based on the following entropy generalized form:
\begin{eqnarray}\label{E:tsallis}
S_q = k\,\frac{1-\sum_{i=1}^W p_i^q}{q-1}, \quad\quad (q\in R; S_1 = S_{BG})
\end{eqnarray}
where $k>0$ is a constant, $W$ is the total number of possibilities of the system and the real number $q$ is an entropic index that characterizes the degree of non-additivity. This expression becomes equivalent to BGS entropy in the limit $q\to1$~\citep{Tsallis1988}. Tsallis entropy is non-additive in such a way that, for a statistically independent system, the entropy of the system is defined by the following pseudo-additive entropic rule:
\begin{eqnarray}\label{E:pseudoadditive}
S_q (A+B) = S_q(A) + S_q(B) + (1-q)S_q(A)S_q(B).
\end{eqnarray}
Consequently, $q = 1$ corresponds to additivity, $q < 1$ corresponds to sub-additivity and $q > 1$ to super-additivity~\citep{Tsekouras2005}. So, considering $S_q\geq0$ in the pseudo-additive formalism of Eqn.~\eqref{E:pseudoadditive}, three entropic classifications are defined:
\begin{align}\label{E:classifications}
\text{Sub-additivity}~(q<1): S_q(A+B) < S_q(A) + S_q(B)\\
\text{Additivity}~(q=1): S_q(A+B) = S_q(A) + S_q(B)\\
\text{Super-additivity}~(q>1): S_q(A+B) > S_q(A) + S_q(B)
\end{align}
The entropic index $q$ varies according to the system properties, which includes the multi-fractal characteristics and long distance correlations, and represents the degree of non-additivity of the system~\citep{Albuquerque2004}.

Similarity voxel measurements provide fully automatic registration techniques for intra and inter-modal medical images using distinct cost functions, which can be determined through different mathematical/statistical approaches. Mutual Information is the most popular and accepted technique in this context and, traditionally uses Shannon's entropy to quantify the information.

Several variations of MI approach have been proposed in order to improve registration reliability and robustness~\citep{Antoine1998,Wachowiak2003a,Wachowiak2003b,Shitong2005,Gao2008,Andronache2008,Cahill2010,Gao2008}. \cite{Studholme1999} proposed the Normalised Mutual Information (NMI) for neuroimage registration and found out that the normalised entropy measure provides significantly improved performance over a range of imaged fields of view. Recently, we have studied the MI using Tsallis entropy (MIT) for intra and inter-modality registration of cerebral MRI and SPECT images. Our previous study using computational simulations found relevant evidences that this approach may provide with contributions to the medical imaging registration scenario~\citep{Amaral-Silva2010}.

Our objective here was to evaluate MI based on Tsallis entropy (MIT and NMIT) as cost functions for different modalities of neuroimage (CT, MRI, PET) registration. For this, we conducted three experiments on current entropy measures (ECC, MIS and NMI) and on the proposed measures (MIT and NMIT) and we compared them as suggested by~\cite{Studholme1999} previously, in the MNI validation. The experiments performed were: 
\begin{enumerate}
\item[1)] The image overlap problem -  Simulation Experiment: to examine the behavior of measures to the misalignment,
\item[2)] Response to varying field of view - Capture Range Experiment: to examine the behavior of the registration algorithm using different measures when it is presented to clinical data with varying fields of view, and 
\item[3)] Accuracy over a range of clinical images - Clinical Experiment: to examine the accuracy of estimates provided by the measures over a lager database of images provided by the Vanderbilt Retrospective Image Registration Evaluation (RIRE) Project.
\end{enumerate}

Our first hypothesis was that MIT and/or NMIT might improve the neuroimage registration of PET to MRI and CT to MRI. A previous study focused on automatic segmentation of cerebral MRI found improvements using Tsallis entropy when compared with Shannon entropy~\citep{Diniz2010}. 

Our second hypothesis was that the non-additive variable q close to value $1.0$ and Shannon entropy will produce similar results, given that Tsallis entropy is a generalization of BGS statistics where this proposed formalism becomes equivalent to Shannon entropy in the limit $q\to1$. We expected to find out optimized $q$ parameters with values different than $1.0$.  
Our alternative hypothesis was that MIT and/or NMIT would be unable to provide a reliable cost function for automatic neuroimage using the MI approach.

\section{Materials and Methods}\label{sec:materials}

\subsection{Datasets}\label{ssec:images}

We used three different set of images:
\begin{enumerate}
\item For the Simulation Experiment: the model used represents alignment of 2D coronal slices through the brain by using half-circled images from different fields of view.

\item For the Capture Range Experiment: MRI (geometrically corrected), CT and PET from a specific patient (patient \#5) of the Vanderbilt study were truncated in-plane at three levels to produce three pairs of images with three different fields of view.

\item For the Clinical Experiment: MRI (not corrected and corrected geometrically), CT and PET from all of the 18 patients provided by the Vanderbilt study~\citep{West1997}.
\end{enumerate}

The Vanderbilt study, also known as Vanderbilt Retrospective Image Registration Evaluation Project (RIRE Project) is a large database consisting of images from 18 patients containing one CT and/or PET image volume and a subset of seven MRI volumes (MP-RAGE; spin-echo T1, PD, T2 and the rectified versions of these three). This project proposes that the investigators use this database to determine a rigid body transformation, based on their retrospective registration method, to align CT to MRI, PET to MRI and MRI to MRI volumes. Afterwards, the proposed method was compared to a gold-standard transformation for the corresponding pair.

\subsection{Registration algorithm}\label{ssec:reg}

A new algorithm was coded in order to perform the registration using the cost functions studied in this project. The algorithm was coded in C++ using predefined libraries and classes from BICG code repository. 

In our algorithm, prior to evaluation and optimization of transformation parameters, all image pairs are first resampled to a common cubic sampling resolution using tri-linear intensity interpolation to increase the sampling rate and Gaussian blurring to reduce the sampling rate as necessary in each dimension. In addition to ensuring equivalent sampling resolution, the images are filtered to provide an equivalent spatial resolution (impulse response). This is important because modalities such as PET may have a very broad impulse response (say 8 mm), whereas other modalities such as MRI have a much narrower response (say 1 mm). We wish to examine the co-occurrence of measurements in the two modalities which occur both at a corresponding location and, from an equivalent region of material (determined by the impulse response). This filtering is achieved using a Gaussian kernel to decrease the spatial resolution of the higher resolution modality to that of the lower resolution modality. This pre-processing also ensures that the marginal entropy of the initially higher resolution modality does not dominate the registration measure. To evaluate each measure between MRI and CT or PET, a discrete histogram estimate of the joint probability distribution is made using 64 intensity bins for each modality. The histogram is formed directly by incrementing the voxel count in the bin corresponding to the MRI and CT or PET values, for every MRI voxel with a corresponding voxel in the other modality. Corresponding values in the CT or PET are estimated using tri-linear interpolation of the neighboring voxel values. The two marginal histograms are calculated from this joint histogram and then the joint and marginal entropies are estimated directly from the counts in these three histograms. In order to recover rigid alignment we need to find the 6 rigid parameters, which provide the optimum value of the registration criteria (maximum of mutual information approaches or the minimum of joint entropy). 

For the proposed experiments a simple multi- resolution hill climbing algorithm was used~\citep{Studholme1997}. All images were further sub-sampled using a Gaussian kernel to form lower resolution versions. This produces a simple Gaussian pyramid where smaller objects are removed as blurring is increased resulting in a smooth function of misalignment between low resolution images at the top of the pyramid. These images containing few voxels can be used to provide an efficient initial registration estimate which is then refined using higher resolution images lower in the pyramid. In these experiments registration was started at an image sampling resolution of $6\times 6 \times 6$ mm with an optimization step size of $6$ mm and terminated at a resolution of $1.5\times 1.5 \times1.5$ mm and step size of $1.50/256$ mm.

\subsection{Image processing}\label{ssec:ip}

\subsubsection{Simulation Experiment}

\begin{figure*}
\centering
	\includegraphics[width=16cm]{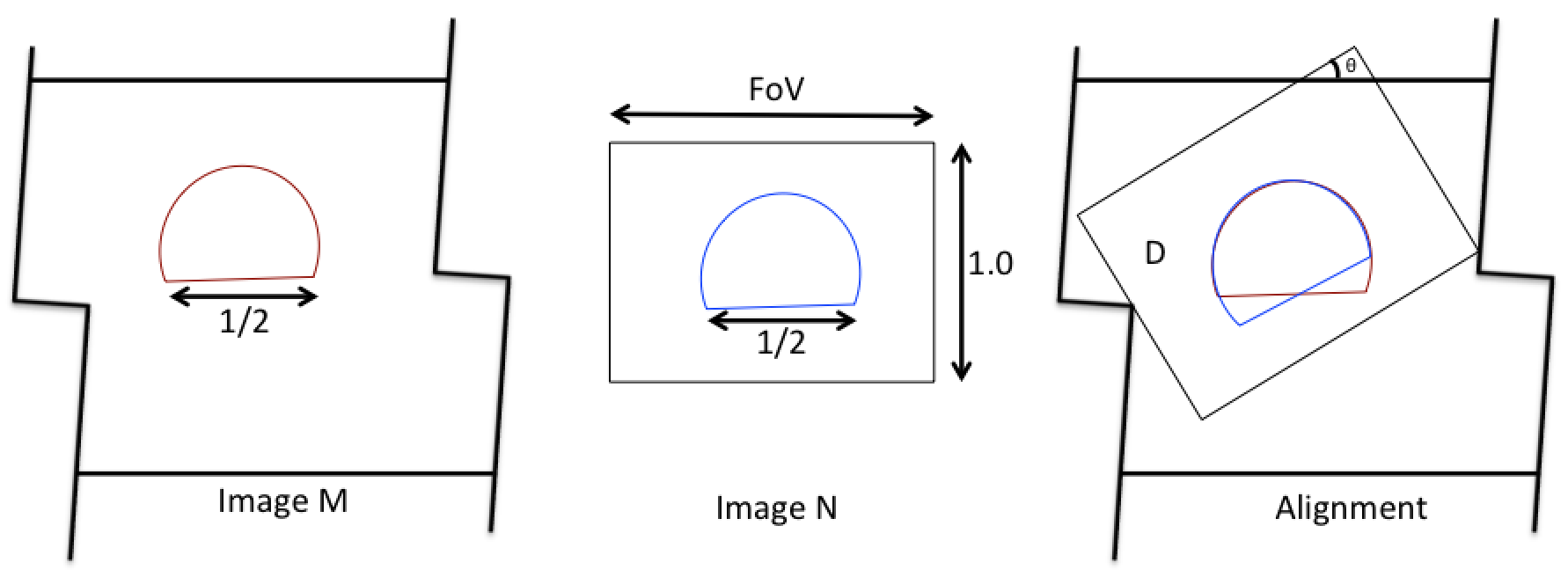}
	\caption{A simple model of rotational alignment $\theta$ between two images of a half circle with varying overlap and horizontal field of view determined by field of view (FOV). Image adapted from~\cite{Studholme1997}.}\label{fig:simple}
\end{figure*}

Initially, we examined the behavior of measurements in the misalignment of an image, in which there were both a variation in overlap statistics and a true point of alignment to be detected by the measure. In this model we had two parameters, one controlling misalignment and one controlling the field of view in which the overlap statistics varied. In order to do this, we used a simple model represented by the registration of 2D coronal slices of brain images of a half-circled image from different fields of view as illustrated in Figure~\ref{fig:simple}.

Horizontally, the first image, for the sake of simplicity, had an infinite extent, while the other had a limited extent determined by the field of view (FOV) parameter. A half circle was placed at the center of the rectangular field of view. We considered in this model the response of different measures to the change in in-plane rotational alignment around the center of the field of view (and circle). To evaluate the measures we calculated the area of overlapping regions in the two images as a function of the rotational alignment and the field-of-view parameter FOV. 

\subsubsection{Capture Range Experiment}

\begin{figure*}
\centering
	\includegraphics[width=16cm]{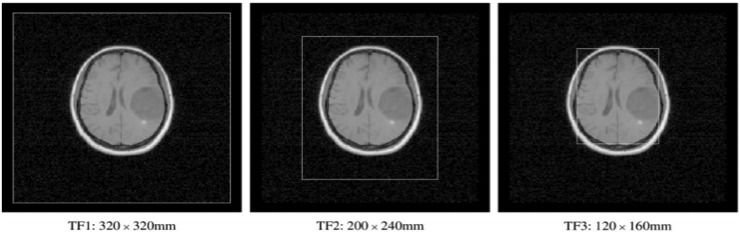}
		\includegraphics[width=16cm]{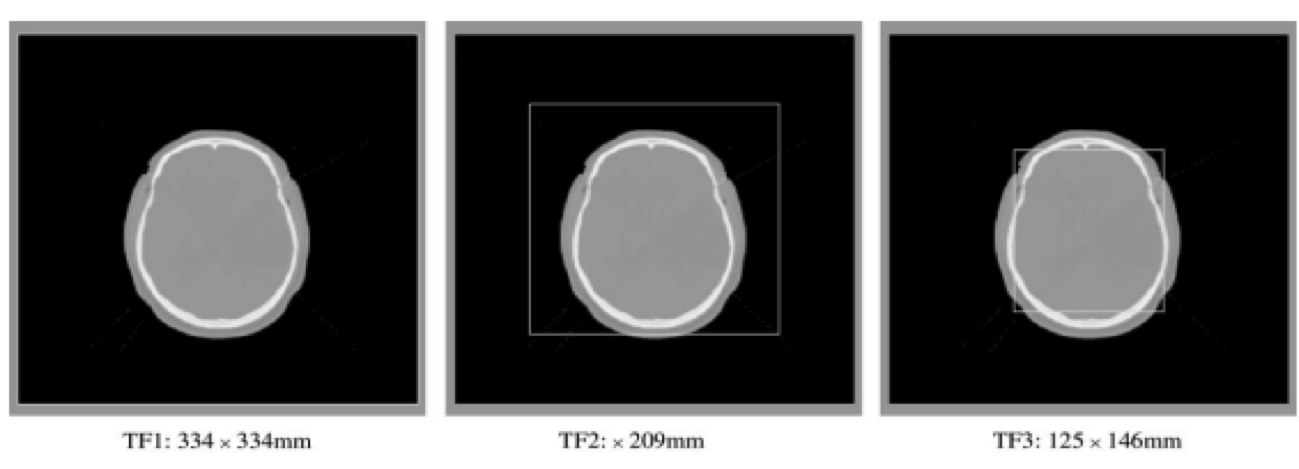}
			\includegraphics[width=16cm]{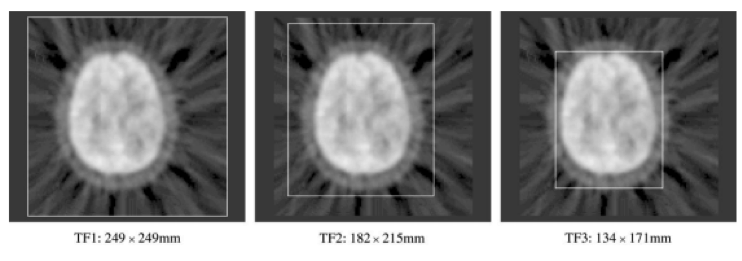}
	\caption{The three levels of (top) MRI, (middle) CT, and (bottom) PET trans-axial field of view used for experiments.}\label{fig:3levels}
\end{figure*}

The aim of these experiments was to examine the behavior of the registration algorithm using different measures when it is presented along with clinical data with varying fields of view. In this case we concentrated on recovering $6$ rigid registration parameters as the trans-axial field of view. Varying the in-plane field of view determines the proportion of the image exhibiting values corresponding to air in the two modalities. In order to look at the effect this has on the registration measures; an image set (patient 5 of the Vanderbilt study) was be truncated in-plane at three levels to produce three pairs of images to register. The fields of view selected (labeled TF1, TF2 and TF3) in the MRI, CT and PET images are shown in Figure~\ref{fig:3levels}.

To examine how robustly different levels of initial misalignment may be recovered using the measures, sets of randomized starting estimates were produced. In these experiments $3$ sets of $50$ randomized transformations were used. These were derived by perturbing the $6$ rigid marker based parameters with random translations and rotations of sizes ($10$mm, $10^{\circ}$),($20$mm, $20^{\circ}$), and ($30$mm, $30^{\circ}$). These were then used as starting estimates for the optimization algorithm in each of the three measurements. 

\subsubsection{Clinical Experiment}

In order to examine the accuracy of estimates provided by the similarity measures over a large database of images, the measures studied were used to recover alignment between all the geometrically corrected image pairs with the original full field of view in the Vanderbilt image database. The starting estimate for the optimization of the measures were the centers of the two image volumes aligned, and no residual rotations between the imaging planes, which represents a typical starting estimate for automated registration in clinical use.

\section{Results and discussion}\label{sec:results}

\subsection{Simulation experiment}\label{ssec:sim}

\begin{figure*}
\centering
	\includegraphics[width=16cm]{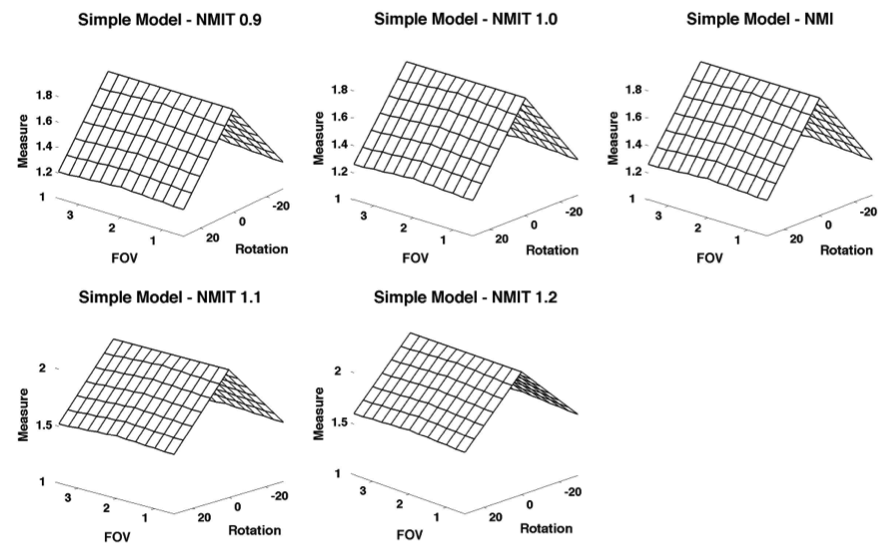}
	\caption{Simulations experiment comparing NNI and NMIT with $q = 0.9$, $\sim1.0$, $1.1$, and $1.2$.}\label{fig:simulations}
\end{figure*}

The simulation experiment showed that for registrations based on the Tsallis entropy with $q$ values close to $1.0$ (and slightly lower and higher) the information measure is maximized as the traditional approaches based on Shannon entropy. It happened for the Mutual Information (MI) and for the Normalized Mutual Information (NMI). See Figure~\ref{fig:simulations}.
 For $q$ higher than $1.0$, the information measure is higher as well, however we observed that it did not meant better alignments. Actually even maximizing the similarity measure, q values higher than $1.0$ performed higher rate of misalignments providing strong evidences that super-additive entropy is not applicable for image registration. It also had been evidenced in previous study with neuroimage registration of brain SPECT and MRI. Posterior experiments confirmed this hypothesis for clinical neuroimage registrations.

\subsection{Experimental design (Capture Range Study)}\label{ssec:exp}

The aim of these experiments was to examine the behavior of the registration algorithm using different measures when it is presented along with clinical data with varying fields of view and different sets of starting points. We did that for both registration modalities: CT to MRI and PET to MRI and then we evaluated the algorithm's performance based on the registration success rate.

\begin{table*}
\centering
\caption{Successfully recovered alignments from $50$ random starts at three levels of misalignments in varied field of views for CT to MRI registration.}\label{tab:CT_MRI}
\begin{tabular}{lccccccccccc}
\hline
	& \multicolumn{3}{c}{$10$mm $10^\circ$}	 &	&	\multicolumn{3}{c}{$20$mm $20^\circ$}	&	&	\multicolumn{3}{c}{$30$mm $30^\circ$}\\
	\cline{2-4}	 \cline{6-8} \cline{10-12}
Measure		&	TF1	&	TF2	&	TF3	&	&	TF1	&	TF2	&	TF3	&	&	TF1	&	TF2	&	TF3\\
Correlation	&	50	&	50	&	50	&	&	50	&	50	&	50	&	&	39	&	47	&	50\\
	\hline
MI			&	50	&	50	&	50	&	&	42	&	50	&	50	&	&	29	&	48	&	50\\
MIT $0.9$		&	50	&	50	&	50	&	&	42	&	50	&	50	&	&	29	&	48	&	50\\
MIT $\sim1.0$	&	50	&	50	&	50	&	&	42	&	50	&	50	&	&	29	&	48	&	50\\
MIT $1.1$		&	34	&	34	&	40	&	&	42	&	45	&	45	&	&	29	&	40	&	43\\
\hline
NMI			&	50	&	50	&	50	&	&	50	&	50	&	50	&	&	50	&	50	&	50\\
NMIT $0.9$	&	50	&	50	&	50	&	&	50	&	50	&	50	&	&	47	&	50	&	50\\
NMIT $\sim1.0$&	50	&	50	&	50	&	&	50	&	50	&	50	&	&	50	&	50	&	50\\
NMIT $1.1	$	&	37	&	39	&	41	&	&	40	&	40	&	46	&	&	29	&	31	&	39\\
	\hline
\end{tabular}	
\end{table*}

According Table~\ref{tab:CT_MRI} we observed that for CT to MRI (T1)  registration, NMI and NMIT obtained $100\%$ success for all sets of transformations and FOV variation proposed. MI obtained $100\%$ success for the first transformation set independently of the FOV, however it was not observed for the second and third transformation sets mostly for high FOV (TF1). Generally, MIT and NMIT using $q$ values lower than $0.8$ and higher than $1.1$ presented high rates of misalignment, on the other hand, those $q$ values close to $1.0$ ($0.9$, $\sim1.0$ and less frequently $1.1$) presented results very similar to the traditional Shannon approaches (MI and NMI), mainly for $q \sim1.0$ (MIT $\sim1.0$ and NMIT $\sim1.0$), which got exactly the same results than the MI and NMI.  NMI had already been proved to be invariant previously in $1999$ by~\cite{Studholme1999} and it has been confirmed in this study which we also evidenced the same characteristic for NMIT $\sim1.0$.

\begin{table*}
\centering
\caption{Successfully recovered alignments from $50$ random starts at three levels of misalignments in varied field of views for PET to MRI registration.}\label{tab:PET_MRI}
\begin{tabular}{lccccccccccc}
\hline
	& \multicolumn{3}{c}{$10$mm $10^\circ$}	 &	&	\multicolumn{3}{c}{$20$mm $20^\circ$}	&	&	\multicolumn{3}{c}{$30$mm $30^\circ$}\\
	\cline{2-4}	 \cline{6-8} \cline{10-12}
Measure		&	TF1	&	TF2	&	TF3	&	&	TF1	&	TF2	&	TF3	&	&	TF1	&	TF2	&	TF3\\
Correlation	&	39	&	50	&	50	&	&	48	&	50	&	50	&	&	43	&	50	&	48\\
	\hline
MI			&	41	&	47	&	50	&	&	33	&	42	&	46	&	&	29	&	48	&	50\\
MIT $0.9$		&	41	&	47	&	50	&	&	33	&	42	&	46	&	&	29	&	48	&	50\\
MIT $\sim1.0$	&	41	&	47	&	50	&	&	33	&	42	&	46	&	&	29	&	48	&	50\\
MIT $1.1$		&	34	&	34	&	40	&	&	42	&	45	&	45	&	&	29	&	40	&	43\\
\hline
NMI			&	50	&	50	&	50	&	&	50	&	50	&	50	&	&	50	&	50	&	48\\
NMIT $0.9$	&	49	&	50	&	50	&	&	46	&	49	&	50	&	&	45	&	45	&	48\\
NMIT $\sim1.0$&	50	&	50	&	50	&	&	50	&	50	&	50	&	&	50	&	50	&	48\\
NMIT $1.1	$	&	17	&	40	&	50	&	&	23	&	39	&	50	&	&	3	&	33	&	48\\
	\hline
\end{tabular}	
\end{table*}

As described previously for CT to MRI registration in Table~\ref{tab:CT_MRI}, Table~\ref{tab:PET_MRI} shows the results for PET to MRI registration where we also observed the best performances by using NMI and NMIT $\sim1.0$. Both NMI and NMIT $\sim1.0$ presented results very close to $100\%$ success for the registrations performed, except for 2 misregistration for the third transformation set ($30$mm and $30^\circ$) and TF3. Correlation presented good results for all transformations mainly for TF2 and TF3.  Curiously ECC presented better success rates than MI. ECC was not studied by~\cite{Studholme1999}, however he also evidenced the inconsistency of the MI for the same experiment, which got considerable mis-registration for TF1.

\begin{figure*}
\centering
	\includegraphics[width=13cm]{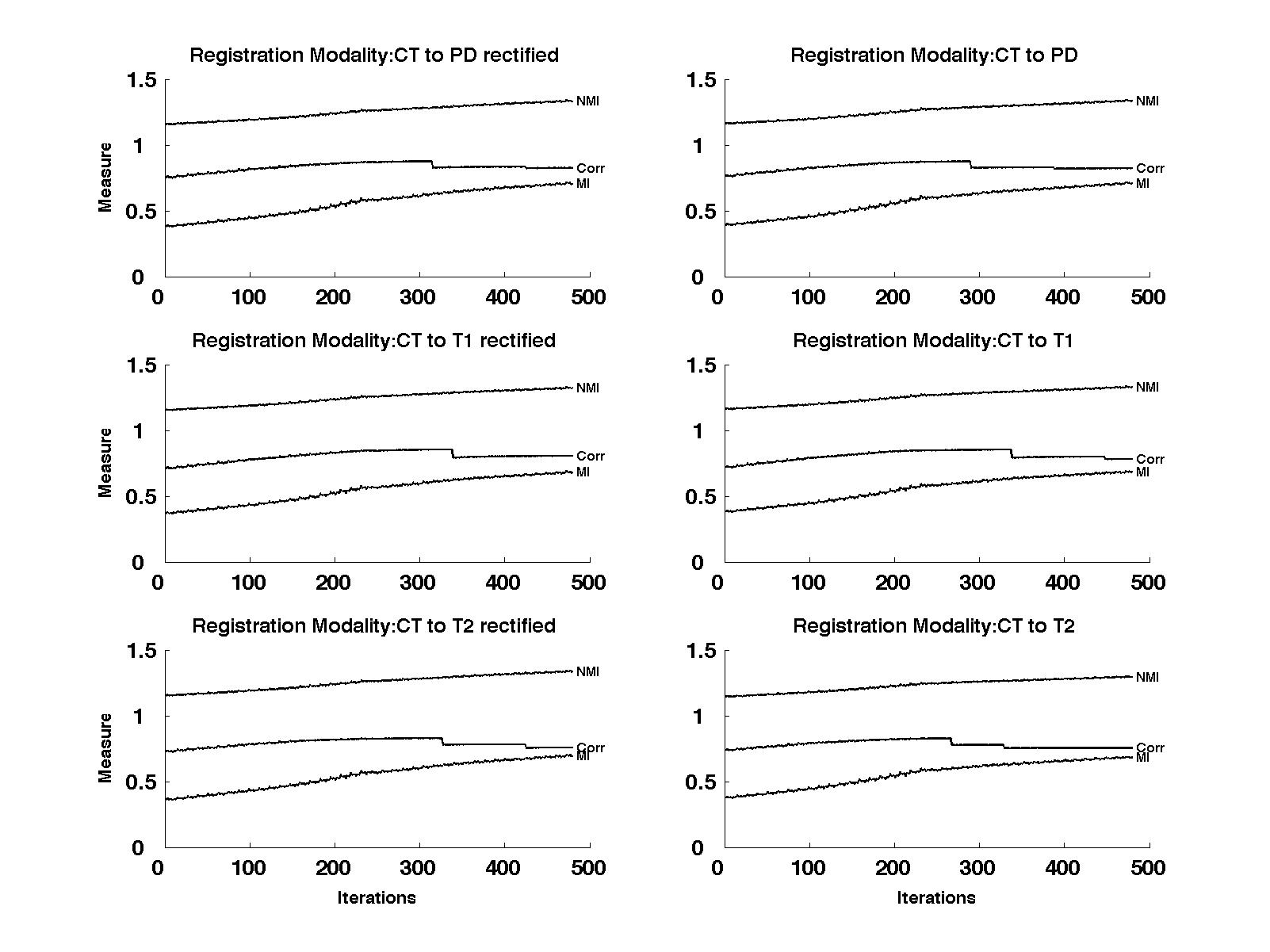}\\
	\includegraphics[width=13cm]{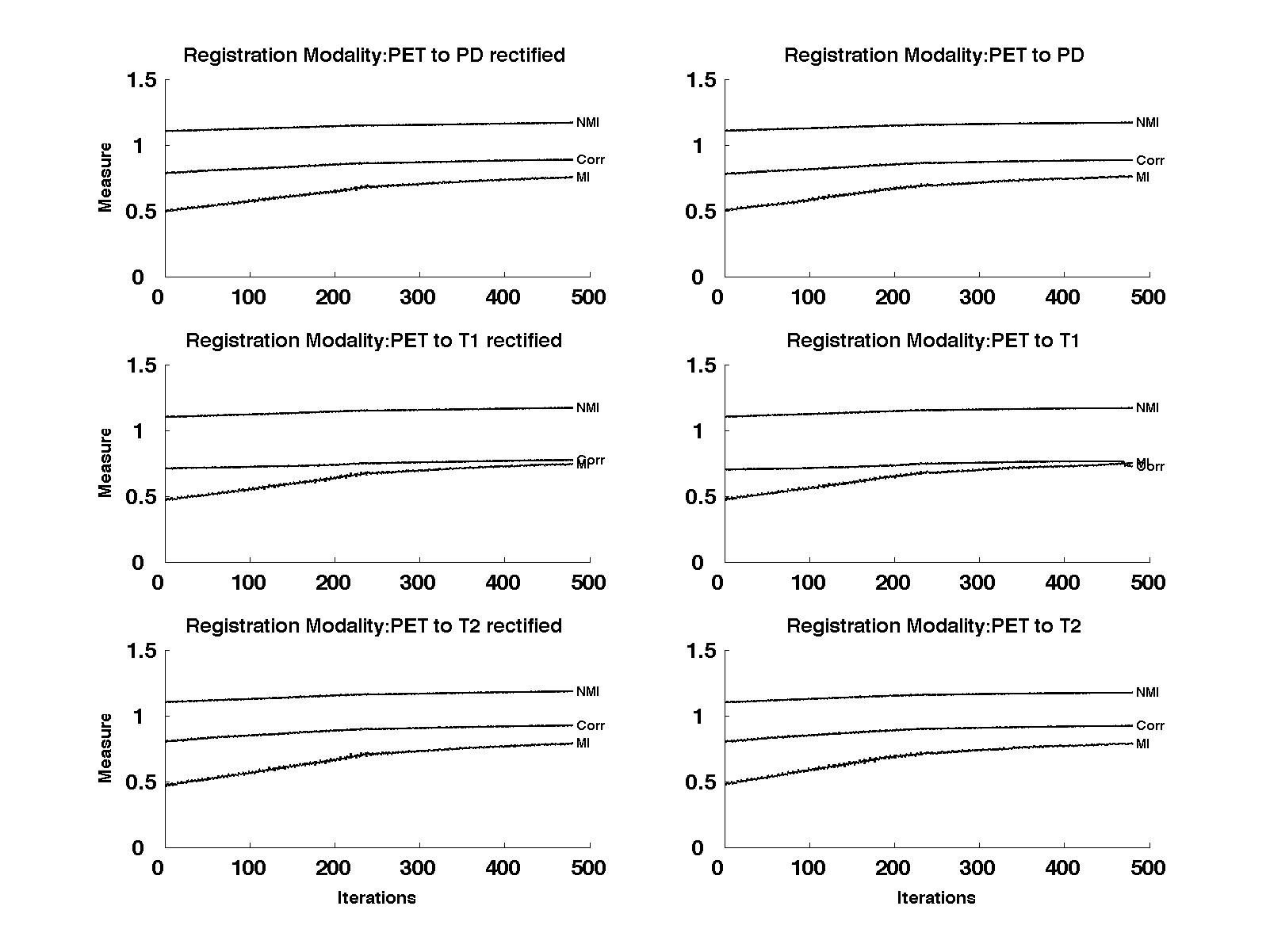}
	\caption{Information measure tracking for the traditional cost functions.}\label{fig:tradional_cost}
\end{figure*}

As observed for CT to MRI registration, for PET to MRI we also noticed that MIT and NMIT using $q$ values lower than $0.8$ and higher than 1.1 presented high rates of misalignment, on the other hand, those q values close to $1$ ($0.1$, $\sim1.0$ and less frequently $1.1$) presented results very similar to the traditional Shannon approaches (MI and NMI), mainly for q $\sim1.0$ (MIT $\sim1.0$ and NMIT $\sim1.0$), which got exactly the same results than the MI and NMI.  Again we evidenced the invariance of the NMI and proved the same characteristic for NMIT$\sim1.0$. For the capture range experiment, we also evidenced the maximization of the similarity measure using NMI for the traditional methods as can be seen in Figure~\ref{fig:tradional_cost}.

\begin{figure*}
\centering
	\includegraphics[width=13cm]{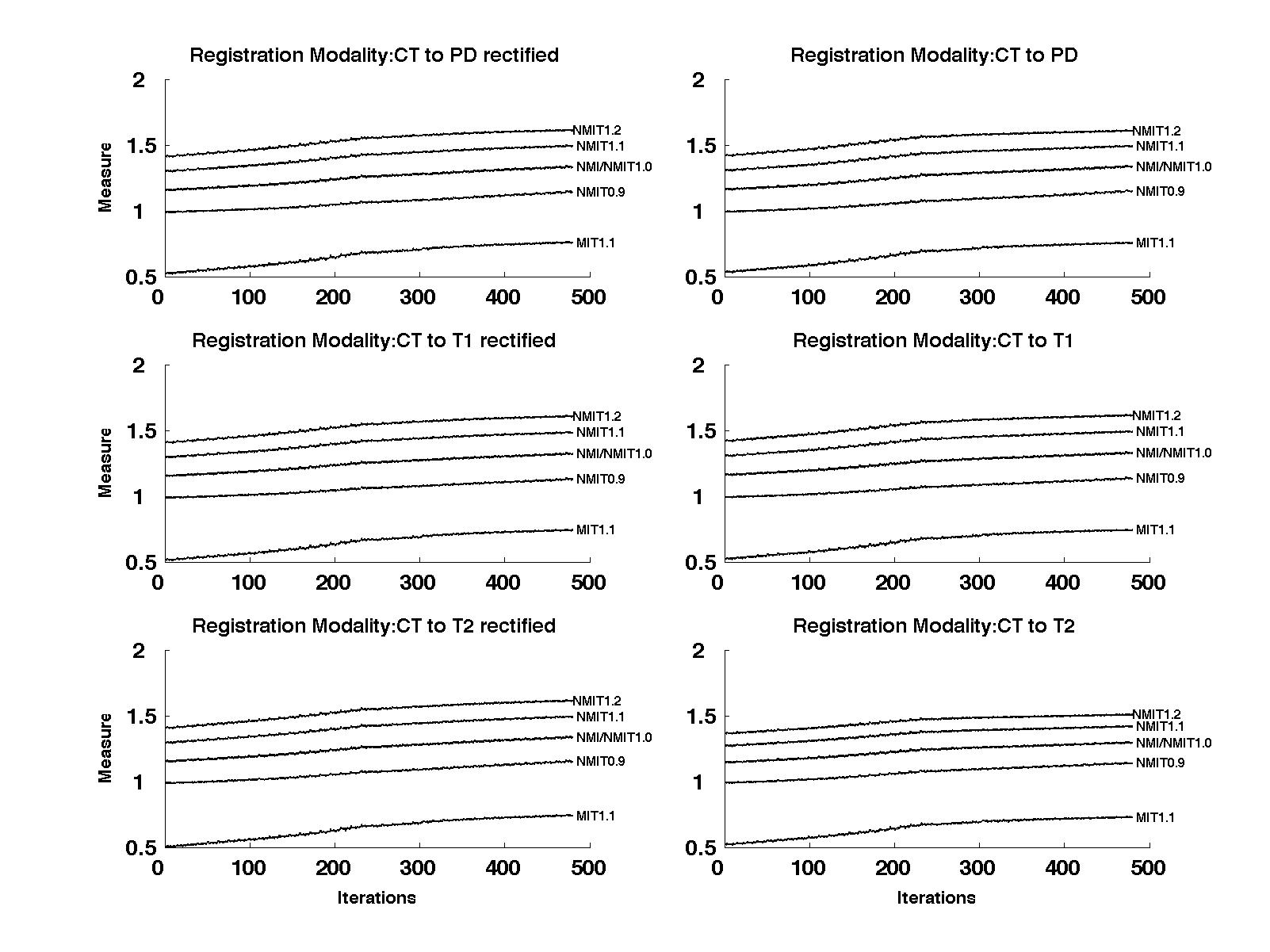}\\
	\includegraphics[width=13cm]{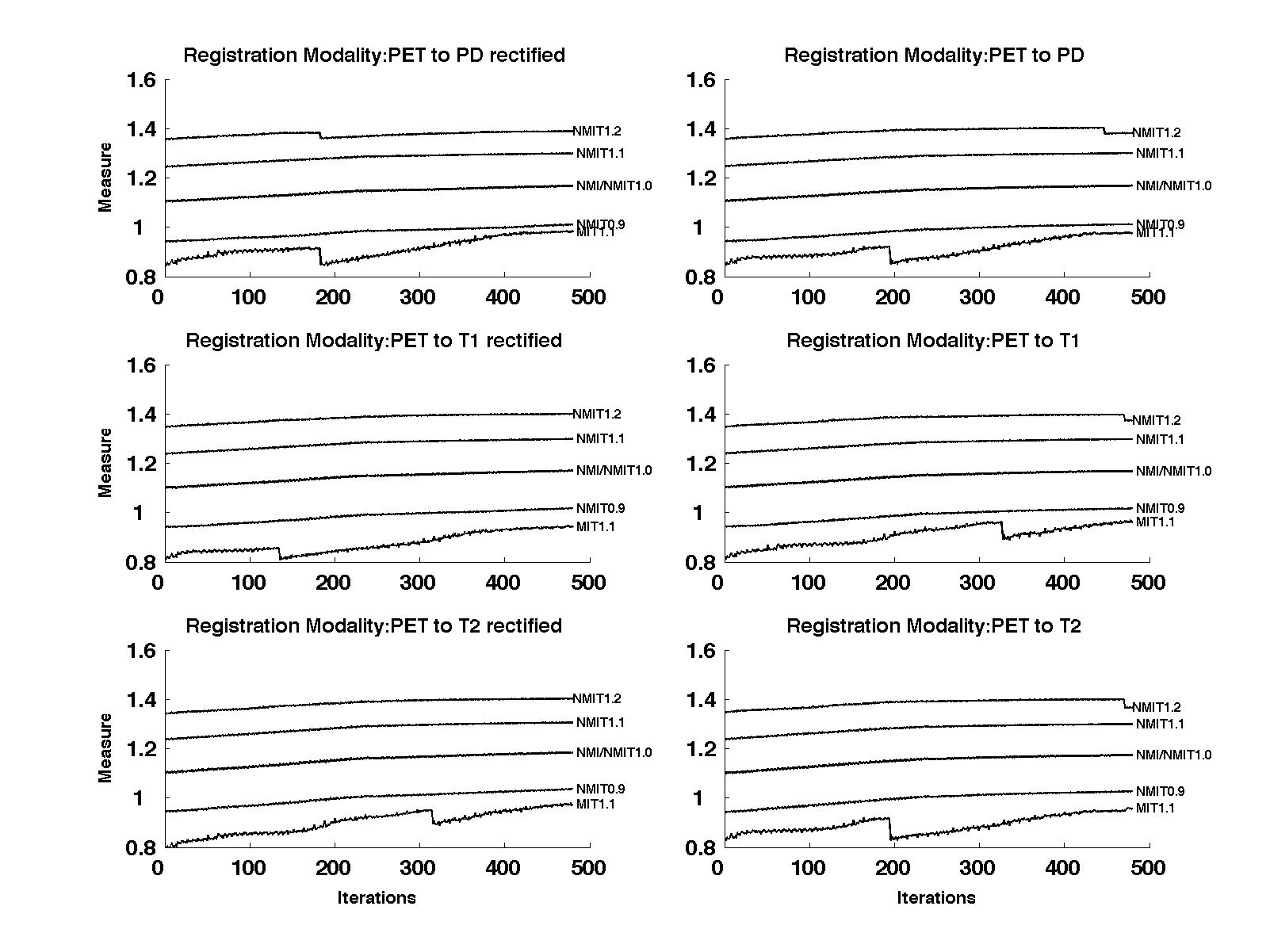}
	\caption{Information measure tracking for the Tsallis based cost functions.}\label{fig:tsallis_cost}
\end{figure*}

For Tsallis entropy we observed that the maximization of the information measure increases according the $q$ value, however, as mentioned before it does not mean the registration gets better results for higher $q$ values; i.e., super-additive entropy maximizes the information measure but does not guarantee reliable registrations. See  Figure~\ref{fig:tsallis_cost}.

\subsection{Clinical Experiment (Accuracy over a range of clinical images)}\label{ssec:clinical}

\begin{figure*}
\centering
	\includegraphics[width=12cm]{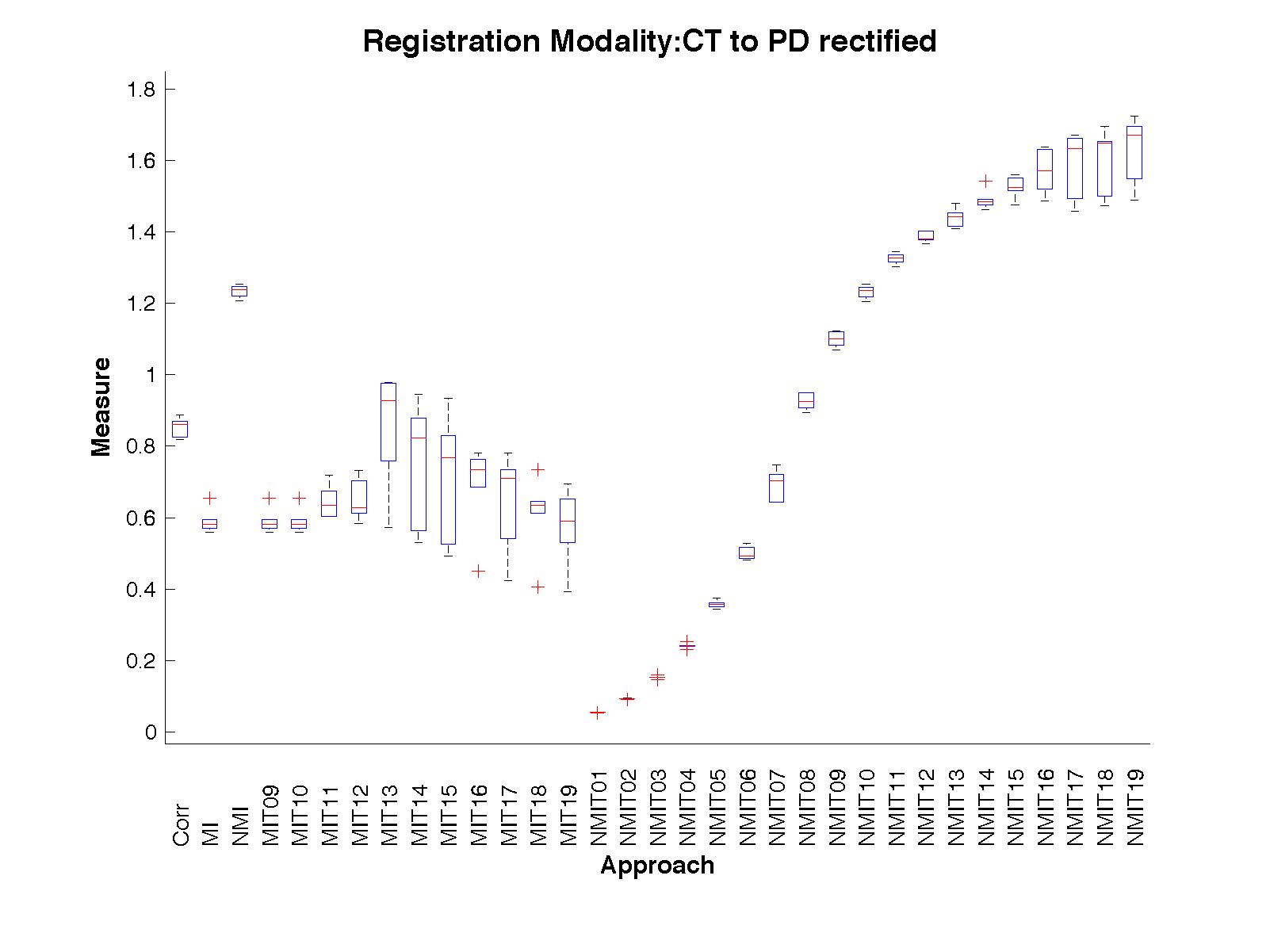}
	\caption{Information measure for CT to MRI PD in clinical experiment (18 subjects).}\label{fig:CTtoPD}
\end{figure*}

\begin{figure*}
\centering
	\includegraphics[width=12cm]{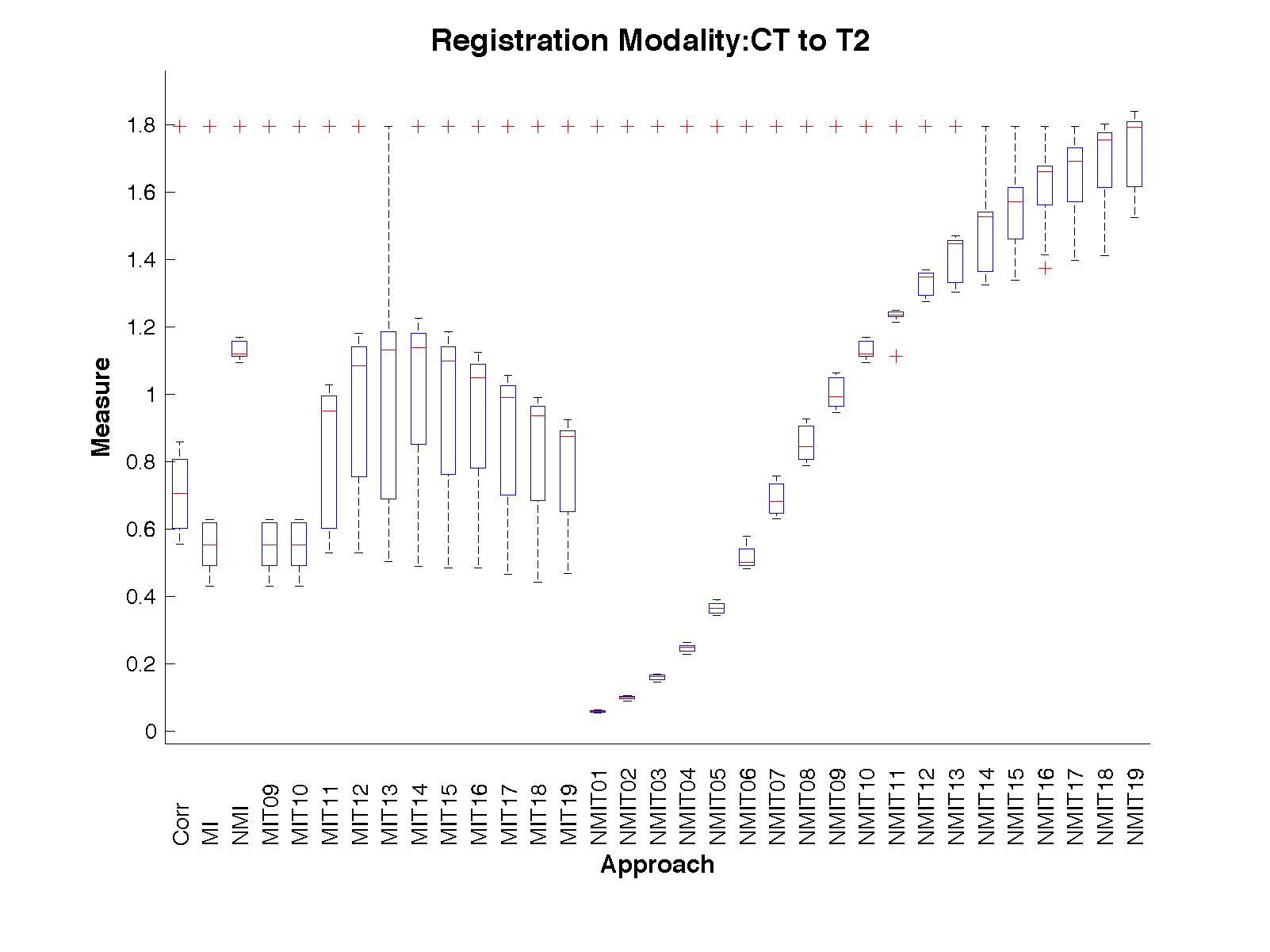}
	\caption{Information measure for CT to MRI T2 in clinical experiment (18 subjects).}\label{fig:CTtoT2}
\end{figure*}

\begin{figure*}
\centering
	\includegraphics[width=12cm]{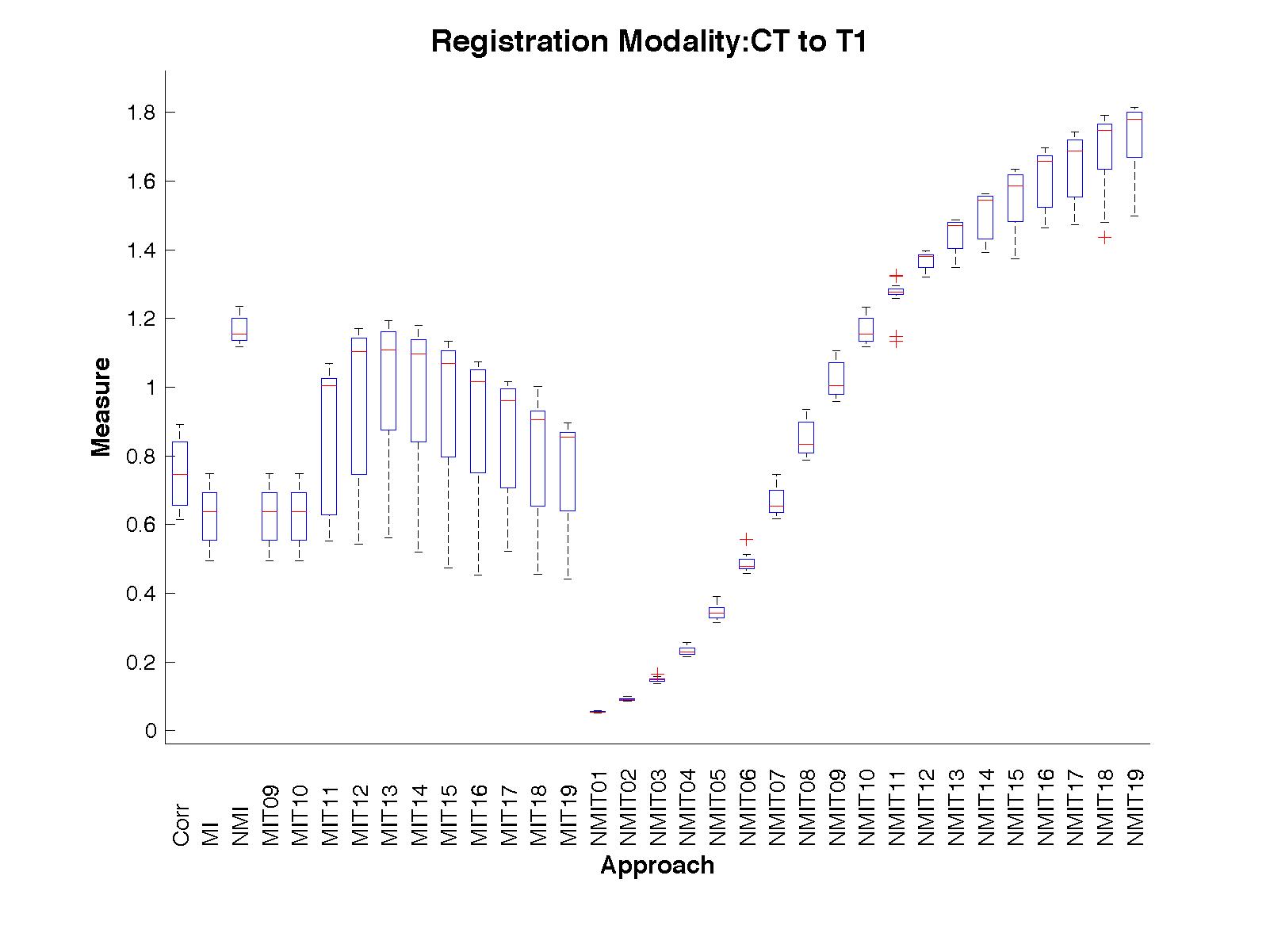}
	\caption{Information measure for CT to MRI T1 in clinical experiment (18 subjects).}\label{fig:CTtoT1}
\end{figure*}

Concerning the clinical experiment, we followed the method proposed by Vanderbilt in the RIRE project.  The transformation is specified by the set of original positions and transformed positions for each of the centers of the voxels at the eight corners of the volume.  Each position is specified by its three coordinates, $x$, $y$, and $z$, in millimeters.  Thus a transformation is specified by $48$ numbers: three numbers for the original position and three for the transformed position for each of eight positions.  We call the coordinates of a transformed position, new\_x, new\_y, and new\_z.  Through the original and the transformed coordinates we obtained the error for each registration submitting the coordinates to evaluation in the RIRE project.

\begin{figure*}
\centering
	\includegraphics[width=14cm]{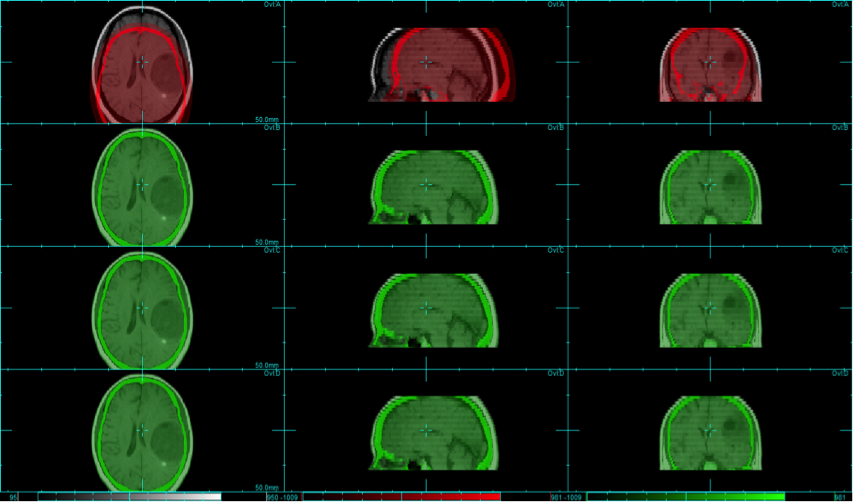}
	\caption{Clinical experiment (patient $\#5$). The first line is presented the overlay of CT with MRI T1 rect. from the patient $\#5$ before registration. The lines $2$, $3$ and for are presented the registration provided by NMI, NMIT 0.9 and NMIT $\sim1.0$. It is possible to observe all of them give satisfactory registrations with imperceptible visual differences among them.}\label{fig:fig11}
\end{figure*}

\begin{figure*}
\centering
	\includegraphics[width=16cm]{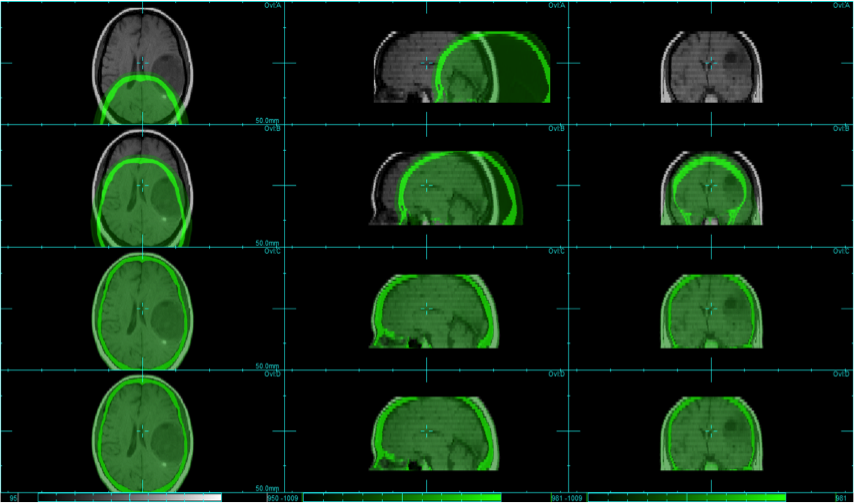}
	\caption{Clinical experiment (patient $\#5$). The first line is presented the overlay of CT with MRI T1 rect. from the patient $\#5$ after registration using NMIT $0.6$; and the second line is by using NMIT $0.7$. As described before, $q \ll 1.0$ are unable to provide satisfactory alignments. Lines $3$, $4$ presented the registration provided by NMI $0.8$ and NMIT $0.9$. They are close to $1.0$ and provide satisfactory registration (by visual analysis), however they present worse registration success rate; which give us evidences of sub-additive formalism does not work properly for neuroimage registration.}\label{fig:fig12}
\end{figure*}

\begin{figure*}
\centering
	\includegraphics[width=16cm]{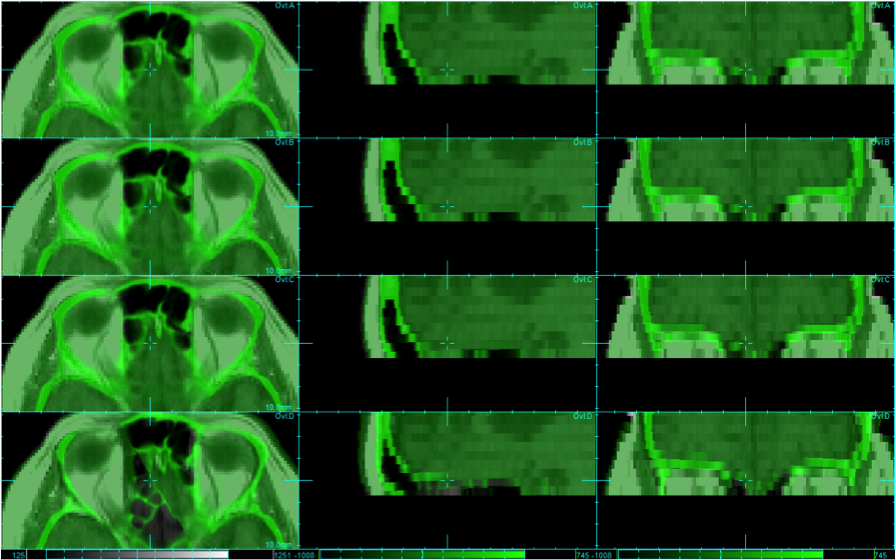}
	\caption{Clinical experiment (patient $\#2$). The three first lines present the overlay of CT to MRI T1 rect. from the patient $\#2$ after registration using NMIT $1.1$, NMIT $1.2$ and NMIT $1.3$ respectively; and the fourth line is the registration by using NMIT $1.4$. As described before, values of $q \gg 1.0$ are unable to provide satisfactory alignments.}\label{fig:fig13}
\end{figure*}

We also analyzed the information measure for each cost function. Figure~\ref{fig:CTtoPD} to Figure~\ref{fig:CTtoT1} show NMI maximizing the information measure with higher values than ECC, MI, NMIT [$0.1$ to $0.9$]. NMI and NMIT $\sim1.0$ get exactly the same information measure value, which is increased as the q value is higher. The same behavior is observed for CT to MRI (PD, T1 and T2) and for PET to MRI (PD, T1 and T2). Below some registrations are presented in Figure~\ref{fig:fig11} to Figure~\ref{fig:fig13}. 

\section{Conclusions}\label{sec:conc}

After executing the experiments and analyzing the results we obtain the following important conclusions:
From Simulation Experiment we concluded that NMIT and MIT maximize the information measure proportionally we increase the non-additive $q$ parameter value, however for $q > 1.0$ the algorithm uses a super-additive entropy formalism which, although maximizes the information measure does not guarantee registration reliability. We also evidenced that $q$ values close to $1.0$ (which defines an additive entropy formalism as NMI and MI) produced better results even thought maximizing information measure in lower proportion than $q \gg 1.0$. Additionally NMIT $\sim1.0$ and MIT $\sim1.0$ produced results similar to NMI and MI; this is a strong evidence that additive entropies seem to work better for neuroimage registration of CT to MRI and PET to MRI.

From the capture range analysis we observed that NMI and its equivalent version from Tsallis entropy (NMIT $\sim1.0$) produced the best success rate for the registrations performed. NMIT $0.9$ had its performance slightly worse than NMIT $\sim1.0$ (and NMI) for CT to MRI and considerably worse for PET to MRI, mainly for the transformation proposed from the set $3$ ($30$mm and $30^\circ$). Values for $q \ll  1.0$ and  $\gg 1.0$ were not able to converge the images and perform the registrations efficiently. From this analysis we can conclude that for both registrations modalities examined (CT to MRI and PET to MRI) $q$ values close to $1.0$ perform registrations with success rates pretty similar to the NMI. 

From the Clinical Experiment, we confirmed the results found out in the Simulation analysis: NMI presented higher information measure maximization than ECC and MI e even higher than MIT $\gg 1.0$ and NMIT $< 1.0$. As expected, NMI and NMIT $\sim1.0$ got exactly the same results and for the NMIT $\gg 1.0$ the information got even more higher maximization but not better alignments in general. As discussed before, we confirmed that super-additive entropy (with $q \gg 1.0$) does not improve the registration although it promotes higher information measure maximization. The same behavior happened for PET to MRI as well.

The registration mean error analysis also does not showed improvements in the alignment using sub-additive ($q \ll 1.0$) or super-additive ($q \gg 1.0$) entropy compared to the additive entropy formalism (NMI, MI, NMIT $\sim1.0$ and MIT $\sim1.0$), however $q$ values close to $1.0$ showed good results mainly when small transformation were required in order to register the images. This analysis left one question to be answered: Could Tsallis entropy using $q$ values close to $1.0$ provide better (related to accuracy and robustness) results for  ``fine" registrations? 

Our final conclusion for this work is that Tsallis entropy using additive entropy can be used for neuroimage registration, however it does not show improvements, just reproduce the same result provide by the traditional NMI and MI. This result is closely related to those found out in the project developed in Brazil focused on SPECT to MRI registration.

\section*{Information Sharing Statement}
Data and software developed in this manuscript are available under upon request from H. T. Amaral-Silva.

\begin{acknowledgements}
We gratefully acknowledge the very helpful participation of physicians, colleagues, and technical staff of the Nuclear Medicine Division, Computation and Math Department and the Image Science and Medical Physics Center. The authors also gratefully acknowledge Dr. Sharmishtaa Seshamani, Dr. Xi Cheng and Dr. Sinchai Tsao (Biomedical Image Computing Group Ð Department of Pediatrics University of Washington) for their very helpful support during the experiments. 
\end{acknowledgements}

\bibliographystyle{spbasic}      
\bibliography{neurorefs}   
\end{document}